# FEATURE SPACE EXPLORATION AS AN ALTERNATIVE FOR DESIGN SPACE EXPLORATION BEYOND THE PARAMETRIC SPACE

TOMAS CABEZON PEDROSO[1] and JINMO RHEE[2] and DARAGH BYRNE[3]
[1,2,3]*Carnegie Mellon University, USA.*
[1]*tcabezon@andrew.cmu.edu, 0000-0002-5483-2676*
[2]*jinmor@andrew.cmu.edu, 0000-0003-4710-7385*
[3]*daraghb@andrew.cmu.edu, 0000-0001-7193-006X*

**Abstract.** This paper compares the parametric design space with a feature space generated by the extraction of design features using deep learning (DL) as an alternative way for design space exploration. In this comparison, the parametric design space is constructed by creating a synthetic dataset of 15.000 elements using a parametric algorithm and reducing its dimensions for visualization. The feature space — reduced-dimensionality vector space of embedded data features — is constructed by training a DL model on the same dataset. We analyze and compare the extracted design features by reducing their dimension and visualizing the results. We demonstrate that parametric design space is narrow in how it describes the design solutions because it is based on the combination of individual parameters. In comparison, we observed that the feature design space can intuitively represent design solutions according to complex parameter relationships. Based on our results, we discuss the potential of translating the features learned by DL models to provide a mechanism for intuitive design exploration space and visualization of possible design solutions.

## 1. Introduction

Parametric modeling has acquired widespread acceptance among creative practitioners as it allows the synthesis of various design options and solutions. Changing the parameters in this modeling process, either manually or randomly, can rapidly create a vast set of design variations (Toulkeridou, 2019). Navigating the resulting *parametric design space* — where the design variants are topologically placed by their parameters — is part of the *design exploration* process — a crucial step in the development of new alternatives and design solutions. Exploration of the parametric design space allows creative practitioners many benefits: to reach satisfying solutions, better define design problems, and understand the opportunities and limitations of the possible solutions.
Despite these benefits, design exploration is laborious within the parametric space and challenged along two fronts: comparison and selection (Fuchkina et al., 2018). Parametric design exploration is an iterative process that focuses on the variation of these individual parameters, rather than on the relationship among them (Yamamoto and Nakakoji, 2005). Hence, comparing one design solution with others by their parameters alone does not always result in a superior solution; for example, the variants generated by the local combination of parameters might not match the design requirements. Moreover, infinite alternative design solutions can be generated by inputting new parameter values. Thus, the parametric design space consists of a huge amount of design variants that cannot be fully or sufficiently explored.

We propose an alternative way to construct and examine the design space, by extracting features from a DL model. By comparing and analyzing how the DL *feature design space* differs from the parametric design space, we illustrate the potential of feature design space for design practitioners during the design exploration process and provide a new way to compare, examine and select the design alternatives based on the exploration of a properly constrained design space.

No previous approach to compare the parametric design space and feature design space as design exploration tools has been found. To demonstrate how the feature space compares to the parametric space, we designed an experiment to construct both a parametric design space and a feature design space using the same dataset. The dataset consists of 15.000 synthetic 3D models produced by a parametric algorithm with five parameters. This parametric design space consists of five axes; each axis corresponds to each of the parameters that are used as inputs of the parametric algorithm. Subsequently, this same dataset is used to train a DL model to compress the data into a feature vector of 128 dimensions. Both the parametric space (five-axes) and the feature space (128 axes) are not directly visualizable due to their high dimensionality. Nevertheless, as visual feedback plays an important role in design exploration (Bradner, Iorio and Davis, 2014), we employ a dimensionality reduction algorithm (t-SNE) to the design space. We are able to illustrate the design exploration space, showing how the data is distributed across both the parametric and feature design spaces.

In the next section, we describe the generation of the dataset, as well as the construction of parametric design space and its visualization. In Section 3., we illustrate how training a DL model resulted in a feature space for design exploration and comparison with the parametric approach. Then, in Section 4., we will compare, contrast, and discuss the characteristics of the DL feature space and the parametric space. (Figure 1.)

FEATURE SPACE EXPLORATION AS AN ALTERNATIVE FOR DESIGN SPACE EXPLORATION BEYOND THE PARAMETRIC SPACE

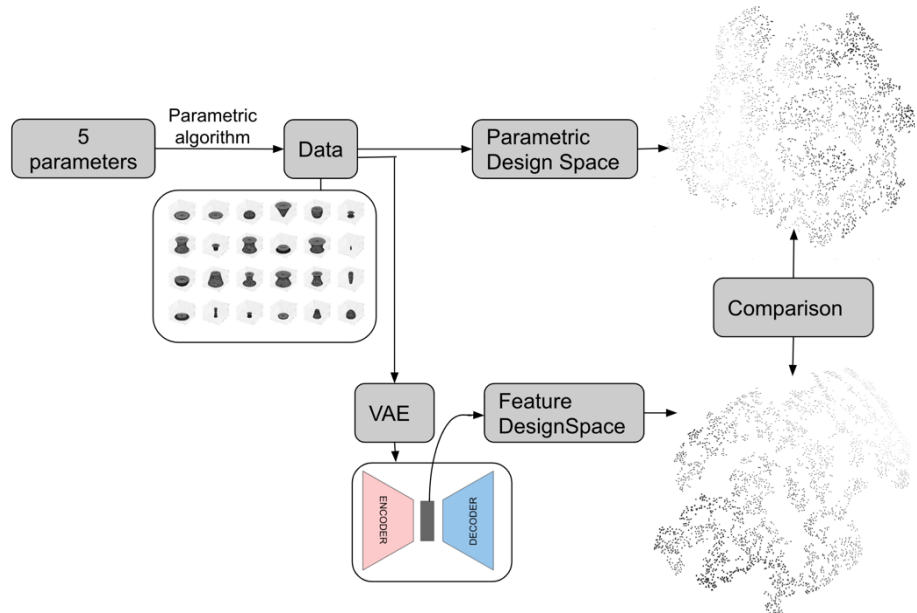

Figure 1. The overall process of comparing parametric design space and feature space from deep learning

## 2. Constructing Parametric Design Space

### 2.1. DATASET GENERATION

To conduct a design space comparison, a simple parametric modeling system was designed: a parametric algorithm for generating different styles of vessels. As with handcraft of pottery wheel throwing, a simple Bezier curve with three control points was turned around an axis to generate each 3D digital vessels; the form of each vessel is specified by the five parameters that were used as inputs. These parameters, as can be seen in Figure 2, are: the height of the vessel, the width of the base, the width of the top opening, and the horizontal and vertical coordinates of the central control point of the Bezier curve that are used to create the curve of the form. The five parameters are represented as a vector, and each vector corresponds to a specific 3D model of a vessel.

Using this system, we created a 3D vessel dataset by randomly generating a total of 15.000 different vessels. The total shape of the parametric representation of the vessel dataset is [15.000, 5], however, as it will be explained in the next section, only 3.000 vessels were used for the space exploration and visualization, so this will be a design space of size [3.000, 5].

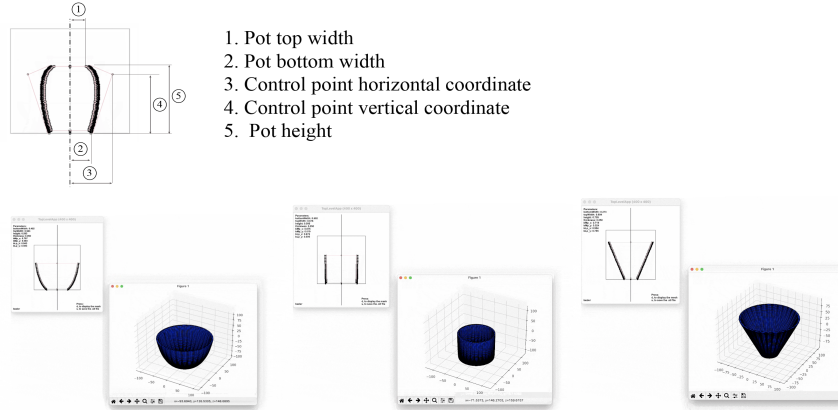

1. Pot top width
2. Pot bottom width
3. Control point horizontal coordinate
4. Control point vertical coordinate
5. Pot height

Figure 2. Upper: An illustration of the dataset parameters. Lower: Three illustrative examples from the dataset with the parameters and the resulting 3D form side-by-side.

## 2.2. DIMENSIONALITY REDUCTION

As a five-dimensional space makes it hard to compare models and to visualize and compare the characteristics, we employed a dimensionality reduction process to reduce the space to two-dimensions and enable the objects to be plotted and compared to one another. Figure 3. shows the overall process of visualizing the space using t-Distributed Stochastic Neighbour Embedding (t-SNE) algorithm (van der Maaten and Hinton, 2008). t-SNE is a popular dimensionality-reduction algorithm for visualizing high-dimensional data. The hyper-parameters used for this reduction are: perplexity: 30; learning rate: 200; and iterations: 1.000.

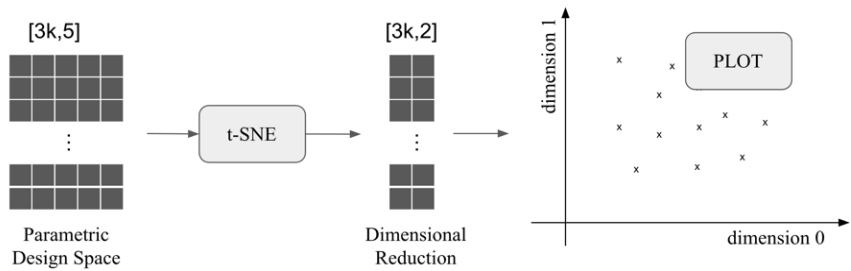

Figure 3. Illustration of the dimensional reduction process for the 3D vessel dataset, and the construction of a parametric design space.

After dimensional reduction, each point in the plot represents the corresponding embedding of a vessel in the parametric design space. Each point is expressed as a 2D image of the profile cut section of the corresponding vessel. Figure 4. represents the reduced parametric design space of the dataset.

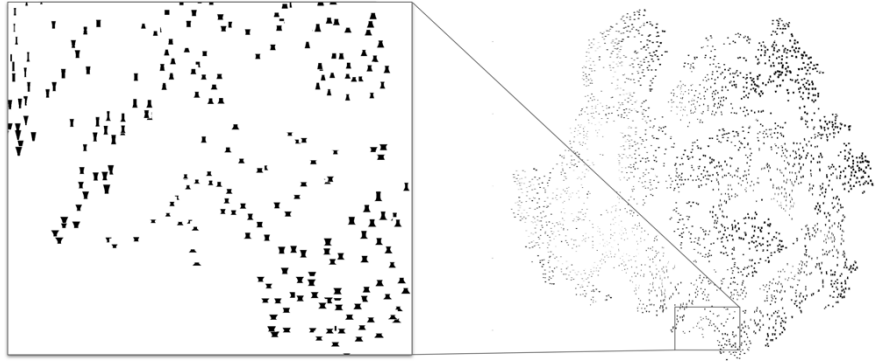

Figure 4. A 2D visualization of the parametric design space of the vessel dataset. Inset image: a detailed section for a subset of the models.

## 3. Constructing the Feature Space

To construct the design space based on the features and not the parameters, we used a Variational Autoencoder (VAE) as a tool for extracting the morphological features of the vessels. VAEs (Kingma and Welling, 2013) are a type of generative deep neural network used for statistical inference problems as they generalize a probabilistic distribution of the given dataset and synthesize new data samples from that distribution. VAEs are composed of two modules: *encoder* and *decoder*. The encoder abstracts the input data into smaller dimensional vectors, latent vectors, and the decoder reconstructs the latent vector back into a 3D shape. During the encoding process, the network captures and extracts the features of the input data. These features can be topologically placed in the data space, namely, *latent space.* In the latent space, the distance between two data points represents the degree of resemblance of data: the closer points, the more resembled. We translate this latent space as the feature space for an alternative way to explore the design space.

### 3.1. DATA PRE-PROCESSING

Different representations of 3D data have been used in DL research, like point clouds (Achlioptas et al., 2018), meshes (Ranjan et al., 2018), or *voxels* (Wu et al., 2017). As resolutions of the data is not key for our purpose rather than the extracted features of it; and because we will implement a VAE for this experiment that needs fixed space inputs for the Convolutional Neural Networks (CNNs), we will be representing our 3D data with voxels. Voxels are discretized three-dimensional grids containing a binary value of volumetric occupancy of an object; they distinguish between the elements on the grid that are filled with material and those that are empty. The size of the voxel will determine the number of divisions of the grid, consequently, the resolution at which we represent our 3D models; the larger size, the more detailed 3D models. In this experiment, we used 32-sized voxels so that a 3D vessel model is represented by 32x32x32 grid, the shape of the entire dataset is [15.000, 32, 32, 32]. Finally, the dataset was then divided into two groups: 80% of the dataset (12.000 vessels) was used for training the DL model, and the remaining 20% (3.000 vessels) was used for testing the model and the parametric and feature space analysis and comparison.

3.2. TRAINING

For training the model, we adopted the VAE architecture implemented in 'Adversarial Generation of Continuous Implicit Shape Representations' (Kleineberg, Fey and Weichert, 2020). The encoder consists of four residual blocks. Each residual block is composed of a 3D convolution layer, followed by a batch normalization and a Leaky ReLu activation layer. The decoder, on the contrary, comprises four residual blocks. Each block starts with a batch normalization, followed by a Leaky ReLu activation layer, and finally a 3D transposed convolution layer. The following hyper-parameters are used for training the VAE with the voxelized vessel dataset: batch size 32, Adam optimizer (Kingma and Ba, 2015), learning rate 5e-. The model was trained in Google Colab Pro using the Nvidia Tesla T4 GPU.

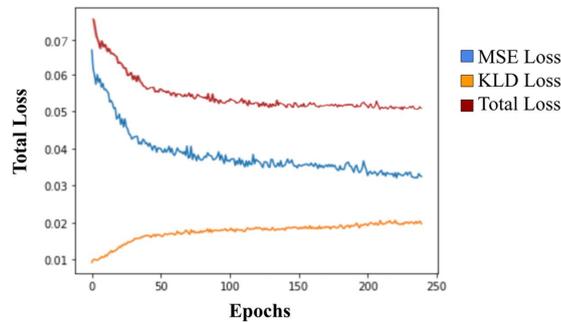

Figure 5. Training process losses.

The model was trained for a total of 240 epochs. We early stopped the model before the model started to overfit, Figure 5. The loss function used during training was a combination of two losses. The first one, is the Kullback–Leibler divergence (KLD) loss (Kullback and Leibler, 1951), with a weight in the total loss formula of 1. This function is a measurement of the difference between two statistical distributions. The second loss is the Minimum Square Distance (MSE) loss (Sammut and Webb, 2010). It is used as the reconstruction loss and measures the error between the input voxels and the reconstructed output. Figure 6. shows the reasonable quality of the reconstruction of the training result after 240 epochs. To ensure the performance of the model, it was evaluated using the test set and showed that the model maintained the accuracy with the new dataset, which shows that the model generalizes well to new data and is able to encode never seen before 3D vessels.

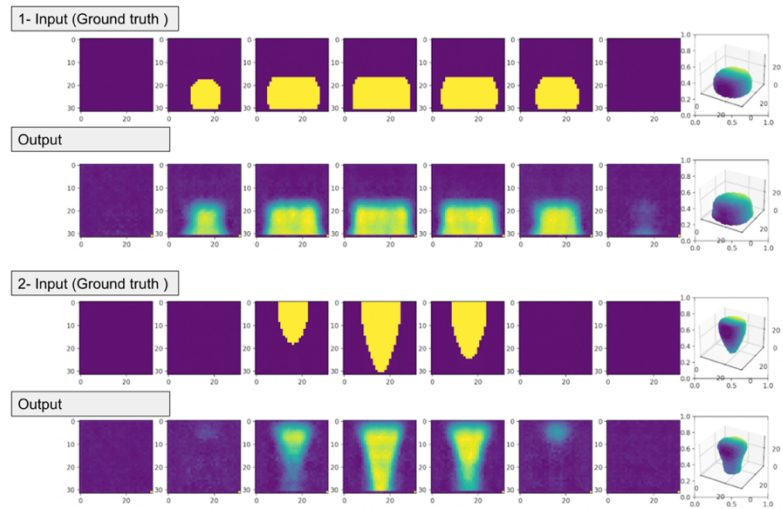

Figure 6. Two examples of reconstructions from the trained VAE: the section slides and 3D voxels of the ground truth (the top row of each example) and the reconstruction (bottom of each example).

## 3.3. DIMENSIONALITY REDUCTION

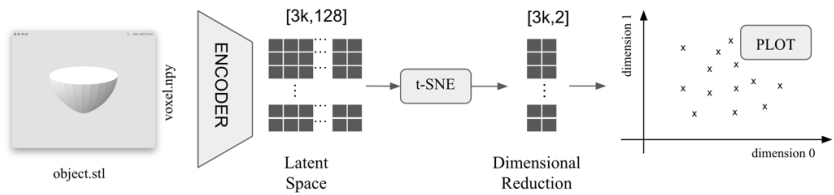

Figure 7. Feature space generation and visualization diagram.

Once the VAE is trained, the encoder is used to extract the features of each vessel in the test dataset from 32.768 dimensions, the size of each voxelized vessel, into 128-dimensional vectors, the latent vectors. Consequently, the entire test dataset of the vessels is represented into vectors whose total shape is [3.000, 128].

Like in the parametric case, 128 dimensions are non-visualizable so the same process as in Section 2 is followed. t-SNE algorithm is used to reduce the dimensionality of each vector and plot the resulting two dimensions in an image with the section of each of the vessels (Figure 7.). The hyper-parameters used for this reduction are: perplexity: 50; learning rate: 700; and iterations: 3. Figure 8. shows the results of distributed feature vectors in the reduced dimensional space, the feature space.

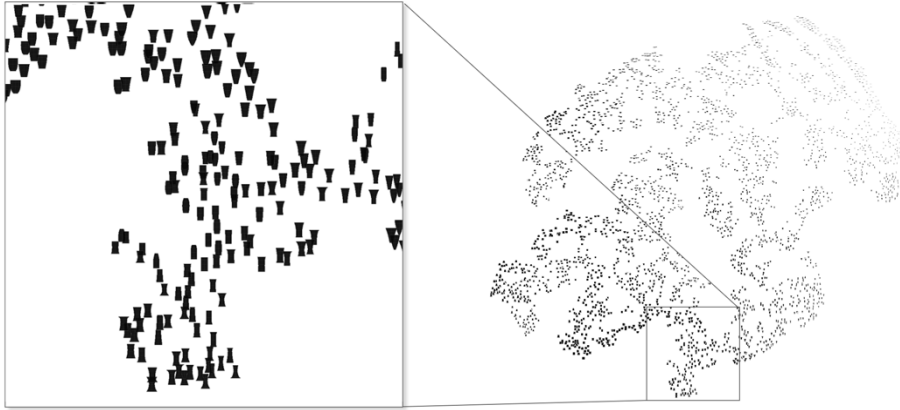

Figure 8. A 2D visualization of the feature design space of the vessel dataset. Inset image: a detailed section for a subset of the models.

## 4. Comparison Between the spaces

Figure 8. shows that similar vessels have been clustered together. Thinner vessels are located at the top right of the image, in contrast to the opposite lower bottom corner with the bigger vessels. The figure illustrates how the VAE model is able to understand the relationship between the parameters and their influence on the output morphological shape.

On the contrary, in the parametric space (Figure 4.), we can see how concave vessels were gathered at the bottom of the image, however, if the height of the vessels is considered, we can see that this parameter was not considered when clustering the vessels. Parametric space is based on each parameter independently, and not on the relationship among them. Therefore, we observe that parametric design space insufficiently expresses the final form characteristics of the vessels by the combinations of the parameters. On the contrary, in Figure 8., the feature space, a gradual change in the shape or concavity as well as height or width is observed.

To further examine and compare the characteristics of both design spaces, we used a clustering, algorithm: a Density-Based Spatial Clustering of Applications with Noise (DBSCAN) (Ester M et al., 1996). It is one of the most common clustering algorithms that finds core samples of high density and expands clusters with them. Figure 9. shows the results of this clustering.

The parametric design space has a total of seven clusters: three of them large, and four of them small. It shows how the parametric design space doesn't provide enough information to intuitively compare the design variants locally, this space shows extreme changes in vessel forms even in the same cluster.

The feature design space, on the contrary, has a total of nine clusters: six main big clusters, and three smaller ones. In the feature design space, we can trace smooth changes in the forms as we move through the different clusters (local changes) and along the whole image (global changes). Shorter vessels are located on the top, while taller ones are on the bottom. If we move on the horizontal axis, the curve that generates the vessels goes from a concave shape on the right to a convex shape on the left. This gives the designer the ability to locally compare similar design alternatives.

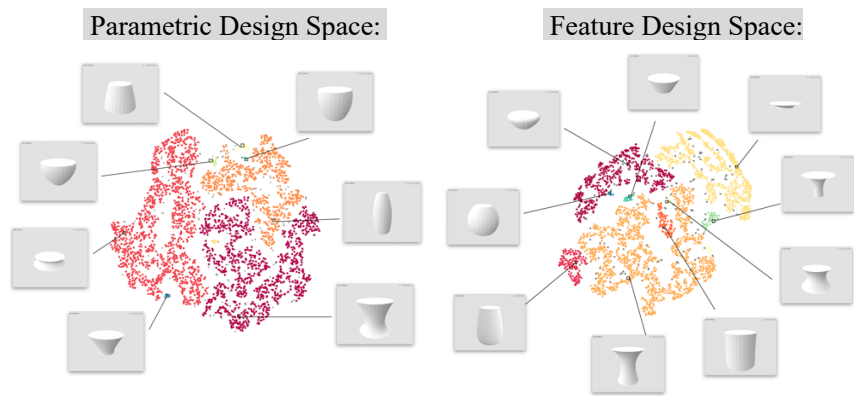

Figure 9. Final visualization and clusters of the parametric and feature design spaces with representative vessels of each group.

## 5. Conclusion and Future work

We constructed the parametric and the feature design spaces using a custom synthetic dataset and a VAE model. By comparing the parametric and feature design spaces, we observed improved distributions of design alternatives in the later. When the multi-dimensional parametric design space is projected into a 2D space (Figures 4. and 9. left), the clusters are insufficiently relevant to the morphological characteristics. On the other hand, when the multi-dimensional feature space is projected into a 2D space (Figures 8. and 9. right), the clusters show sufficient relevance to the features of the data they represent.

Based on this comparison, we conclude that combination of individual parameters in the parametric design space is limited in representing the morphological characteristics of the shapes. However, we showed that DL models can be used to extract design features from 3D models and that the extracted features are more complex than the combinations of individual parameters. Hence, we conclude that the extracted features, that include information of the relationships between the parameters, can construct a well-distributed design space. For that reason, we propose feature design space as a tool for design space exploration that creative practitioners can use as a new way for looking at objects beyond the parametric design space.

Our results and implications are limited to a single dataset and DL model, however the results seem promising. Future work will expand on this study with more diverse datasets generated by more complex parametric algorithms. Accordingly, to perform the feature extraction, we would like to train other types of DL models to investigate different potentials of DL in design.